\documentclass[letterpaper, 10 pt, conference]{ieeeconf}  

\IEEEoverridecommandlockouts                              

\usepackage{graphics} 
\usepackage{epsfig} 
\usepackage{mathptmx} 
\usepackage{times} 
\usepackage{amsmath} 
\usepackage{amssymb}  
\usepackage{algorithm}
\usepackage{algorithmic}

\usepackage{xpatch}
\makeatletter
\patchcmd\@makecaption{\\}{.~}{}{\fail}
\makeatletter

\title{\LARGE \bf
Mesh-based Tools to Analyze Deep Reinforcement Learning Policies\\ for Underactuated Biped Locomotion
}

\author{Nihar Talele and Katie Byl
\thanks{*This work was funded in part by NSF NRI award 1526424.}
\thanks{Nihar Talele and Katie Byl are with the Electrical Engineering Department at the University of California, Santa Barbara CA 93106
        {\tt\small nihar@ucsb.edu},
        {\tt\small katiebyl@ucsb.edu}}
}

\begin{document}

\maketitle
\thispagestyle{empty}
\pagestyle{empty}

\begin{abstract}
In this paper, we present a mesh-based approach to analyze stability and robustness of the policies obtained via deep reinforcement learning for various biped gaits of a five-link planar model. Intuitively, one would expect that including perturbations and/or other types of noise during training would likely result in more robustness of the resulting control policy.  However, one would also like to have a quantitative and computationally-efficient means of evaluating the degree to which this might be so. Rather than relying on Monte Carlo simulations, which can become quite computationally burdensome in quantifying performance metrics, our goal is to provide more sophisticated tools to assess robustness properties of such policies. Our work is motivated by the twin hypotheses that contraction of dynamics, when achievable, can simplify the required complexity of a control policy and that control policies obtained via deep learning may therefore exhibit tendency to contract to lower-dimensional manifolds within the full state space, as a result. The tractability of our mesh-based tools in this work provides some evidence that this may be so.


\end{abstract}

\section{INTRODUCTION}
Although legged locomotion is less energy efficient than wheeled locomotion on relatively mild terrain, it offers the advantage of more flexibility in the variety of terrain that can be traversed. However, this flexibility often comes with added complexity of control, particularly as legged locomotion can involve phases of underactuation. Two approaches have become prominent in recent times for the control of legged locomotion. Model-based trajectory optimization has shown impressive results, for example in its application within the DARPA Robotics Challenge (DRC)~\cite{kuindersma2016optimization,Feng2015OptimizationbasedFB}. Also, with the advent of improved computational capabilities, the field of deep reinforcement learning (DRL) is now being successfully applied to generate control policies for complicated dynamical systems like humanoids~\cite{heess2017emergence}.

Model-based trajectory optimization methods, as described in \cite{7487270,hereid20163d}, involve the generation of a trajectory for the system that optimizes a certain cost function, such as energy minimization. This optimal solution is then used as a reference trajectory for the actual system through the use of a low level controller. The low level controller enforces contraction of the system onto a low dimensional manifold; e.g., for the underactuated dynamics, some subset of directly-controlled degrees of freedom (DOFs) converge rapidly to desired trajectories, compared to other (passive, slower, but still stable) DOF(s). 

This contraction onto lower dimensional manifolds has quite some significance, as it can correspondingly allow for the avoidance of the classic ``curse of dimensionality'' in implementing discrete approximations to analyze the resulting nonlinear dynamics. In the present work, we hypothesize that such contraction may be likely, if not required, also to exist for policies that are (somewhat more mysteriously) derived from deep learning algorithms. Our reasoning is simply that it would be impossible to explore, let alone identify good control actions across, any non-trivial volume of a moderate or large dimensional state space. This implies that any seemingly well-behaved control policy for such systems displays a ``contraction'' behavior, in which the policy shepherds the dynamics to remain in lower-dimensional regions of state space. In this work, we identify and
exploit such contraction, in turn applying meshing tools to evaluate the performance of control policies for such systems. 

In past work within our own group, we have used model-based trajectory optimizations to corroborate human data studies with model-based energy-optimal gaits~\cite{Sebastian2016} and to explore non-intuitive motions for underactuated rolling locomotion systems~\cite{Guillaume}. While relationships between the cost function, constraints in the optimization, low level controller, and the robustness of the obtained policy often have some intuitive characteristics, it is not so clear what quantitative effects the choice of reward function has on the robustness of the obtained policy. With this long-term goal in mind, and with application to quantification of deep learning policies as a particular aim, we analyze locomotion control polices obtained via deep reinforcement learning and propose the use of meshing tools to quantify stability and robustness in terms of failure rates.

Our quantification of failure rates here builds on a body of work in which the long-term dynamics and robustness of walking systems are approximated by creating a discrete mesh of Poincar\'e-based snapshots of the reachable states and possible state-action transitions, for various noisy situations.
In \cite{Byl2009MetastableWM}, Byl and Tedrake propose the mean first passage time (MFPT), measured in terms of average steps-until-failure (i.e., until falling or otherwise coming to rest), as a metric to evaluate the robustness of a rimless wheel and of the compass gait walker, each on rough terrain. In this framework, robust walking is viewed as a metastable dynamic process: capable of very long periods in near-limit-cycle behavior but ultimately guaranteed to eventually fall. Saglam and Byl build upon this work by developing a reachability-based approach to non-uniform meshing~\cite{saglam2014RSS,saglam2015meshing}. In all of these works, the only noise considered is variability in terrain height, i.e., walking on rough terrain. In \cite{Meshing_nihar_2019}, Talele and Byl demonstrate that these mesh-based tools can also be extended to additional noise cases, including push disturbances throughout the gait cycle. In our current work, we expand the applicability of our meshing tools by first showing that it is possible to use the meshing analysis on policies generated via deep reinforcement learning and then use these tools to perform a systematic analysis on the generated policies to demonstrate how these tools can be used to quantify long-term performance of a dynamic system controlled by a policy derived via deep learning.

The rest of the paper is organized as follows. In Section \ref{sec:model}, we present the 5-link model created in MuJoCo \cite{mujoco} that we use for our simulation as well as some information on the deep reinforcement learning policies that we generate. In Section \ref{sec:mesh}, we provide details on the basic meshing concepts that we use for our analyses in the results section. Section \ref{sec:results} presents our results, where we evaluate the robustness of the policies obtained via deep reinforcement learning for different reward functions, following which we conclude in Section \ref{sec:conclusion} by discussing and summarizing the ideas presented along with the possible future directions in which we want to extend the current framework.   

\section{MODEL AND CONTROL}
\label{sec:model}
\subsection{Model}
\begin{figure}[!ht]
\centering
\includegraphics[width = 1.5 in]{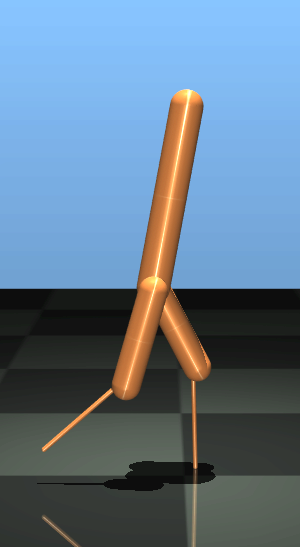}~~~~
\includegraphics[width = 1.5 in]{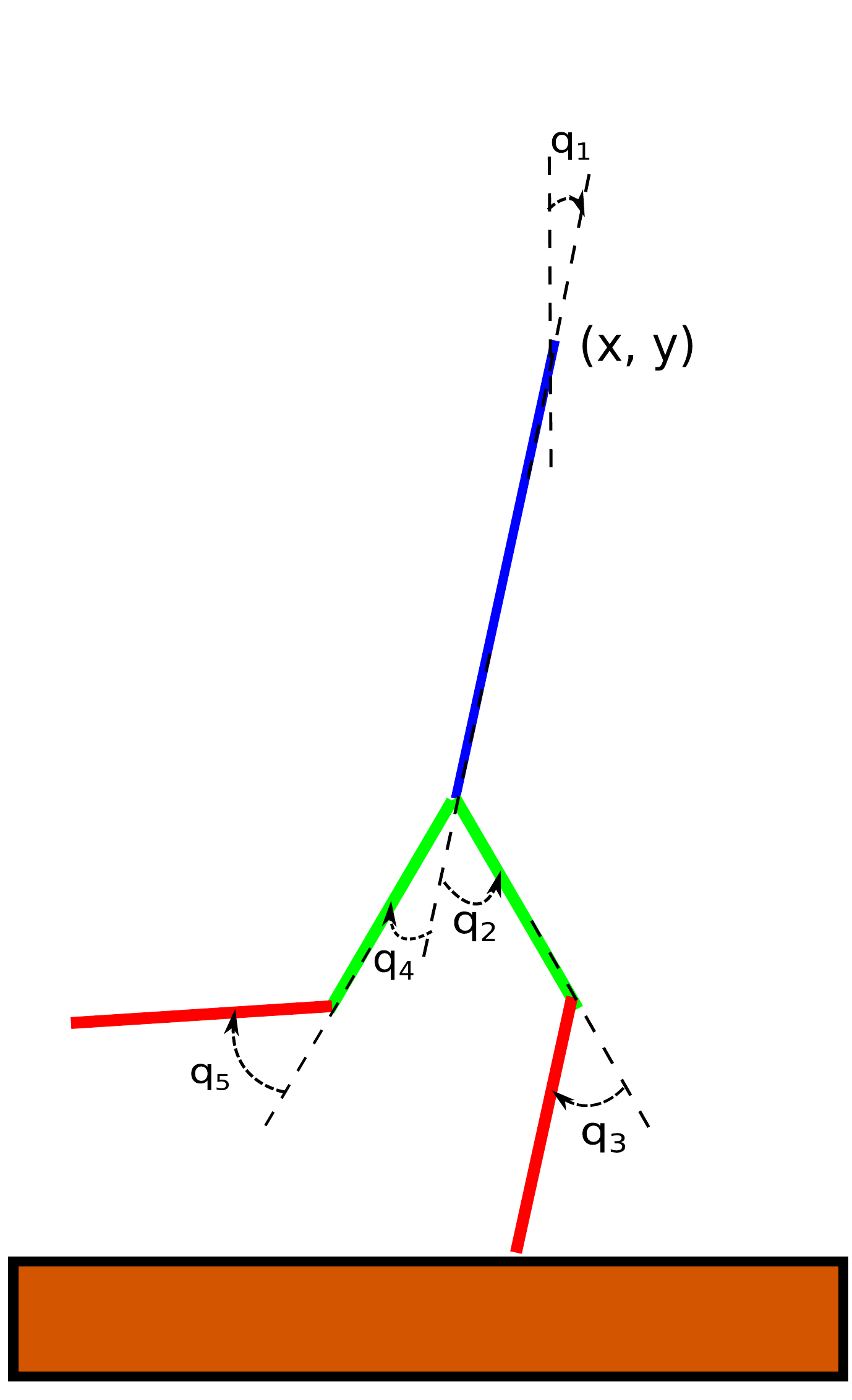}
\caption{The 5-link biped model used in simulations. At left, the planar model in MuJoCo's simulation engine, used for our simulations. The stick figure on the right shows the degrees of freedom of the model on the left.}
\label{fig:model_5link}
\end{figure}

We use a 5-link planar biped model with point feet as shown in Fig.~\ref{fig:model_5link} for our analyses. All simulations for the results presented in this paper are done in MuJoCo~\cite{mujoco}. The model has a torso, two hips and two knees. Its total mass is 70 (kg), and it stands 1.6 (m) tall when fully upright. Length, mass and inertia parameters, which are chosen to approximate a typical human, are identical to those listed for ``Set 5'' within~\cite{Methods_Nihar}. The model has a total of 4 actuators (2 at hips and 2 at the knees) and 7 degrees of freedom $q:=[x,~y,~q_1,~q_2,~q_3,~q_4,~q_5]^T$ where $[x,~y]$ are the position coordinates of the topmost point of the torso and $[q_1,~q_2,~q_3,~q_4,~q_5]$ correspond to the five angles shown in the Fig.~\ref{fig:model_5link}. We ignore rolling friction and set the friction model to oppose sliding in the tangential plane of two contacting bodies only. We set all the viscous damping coefficients to zero. All the contacts in the simulation are soft contacts: MuJoCo models the interaction between two bodies as a soft contact, for efficient computation. We set the integration method to Runge-Kutta (RK4) with a time step of 0.002~(s). All the torque inputs are restricted to $\pm$~100~(N$\cdot$m).

\subsection{Control Policy}
We use the Proximal Policy Optimization algorithm \cite{schulman2017proximal} in the openAI's baselines package \cite{baselines} to train our model and obtain the control policies. The training algorithm models continuous time action space as a probability density distribution that it learns for each observation or state of the environment. While training, the algorithm samples from this distribution and then takes an action for the corresponding observation. During evaluation of a trained policy, however, instead of sampling from the learned distribution, we pick the action that has the maximum likelihood. Each control action is held at a constant value for a total of 4 consecutive time steps during simulation. Thus, even though the time step for the integration scheme we use is~0.002~(s), the action sampled for the current observation or the environment state is applied for a total of 0.008~(s). We set the maximum training episode length to 4000 time steps, and each state observation is clipped to a maximum magnitude of 10.

\section{MESHING}
\label{sec:mesh}
We create our mesh $\mathbb{M}$ by deterministically mapping the reachable state space of the system on a Poincar\'e section for various disturbances, $\gamma$, which for this work are pushes of a variety of magnitudes and timing within the gait cycle. Because there are no constraints requiring left-right symmetry of the locomotion policy learned, we perform a Poincar\'e analysis on a full gait ``cycle'', i.e., after two steps are taken. A Poincar\'e section is taken when the left foot (an arbitrary choice) makes an impact with the ground. We denote the post-impact state as $s = x^+$. The mesh has one self-absorbing state (state \#1) to which all failure events transition. In general, our meshing tools can allow us to calculate the optimal (switching) policy from a set of controllers, but for the current work, we simply analyze the policies individually obtained via DRL framework as explained in Section~\ref{sec:model}. Once the left foot impacts the ground, we compare the post-impact state to previously observed post-impact states using the following metric: 
\begin{equation}
\label{eq:distance_metric}
d(s_i)=\min_{s_{j} \in \mathbb{M}}\sqrt[]{\sum_{k=1}^{n}(s_i(k)-s_j(k))^2}.
\end{equation}
If $d(s_i) > d_{tr}$, where $d_{tr}$ is some distance threshold, the state $s_i$ is added to the mesh. 
A deterministic state transition matrix $T_{det}(s,\xi,\gamma)$ which describes all state transitions is maintained and updated at every iteration, where $\xi$ is the particular DRL policy being analyzed. In case of failure, the corresponding transition goes to state \#1, indicating that under the current control policy, the state has transitioned to the absorbing failure state in the mesh. 

Our disturbance set consists of various pushes $\gamma$ happening at different times in forward as well as backward directions. Each push occurs at the center of mass (COM) of the torso link. The disturbance profile is characterized by a certain probability distribution that we chose, and it is denoted by $P(\gamma)$. 
Algorithm~\ref{alg:meshing} outlines our meshing procedure.

\begin{algorithm}
\caption{Meshing Algorithm}
\label{alg:meshing}
\begin{algorithmic}
\REQUIRE Initial matrix of states S, containing two states: $s_1$ (failure) and $s_2$ (one initial, non-failure state). \\
Disturbance profile D. \\
Distance threshold $d_{tr}$.\\
Set of controllers Z.

\ENSURE Final matrix of all states S, discretely spanning the reachable state space. \\
Deterministic state transition matrix $T_{det}(s,\xi,\gamma)$.  \\
\vskip 0.1in
\STATE \textbf{Initialization:}\\
\STATE Current state index: cur = 2
\STATE Total number of states: nstate = 2
\WHILE{cur $\leq$ nstate}
\STATE xcur $\leftarrow$ S(cur) \\
\FORALL{controllers $\xi$ in Z}
\FORALL{disturbances $\gamma$ in D}
\STATE x, flag $\leftarrow$ simulatedynamics(xcur, $\xi$, $\gamma$)
\IF{flag = 0 (Step taken successfully)}
\IF{d(x, s) $>$ $d_{tr}$ $\forall$ s $\in$ S}
\STATE add x to S and set $T_{det}(xcur, \xi, \gamma)$ = nstate \\
\STATE nstate $\leftarrow$ nstate + 1
\ELSE
\STATE set $T_{det}(xcur, \xi, \gamma)$ = index of s for d(xcur, s) $<$ $d_{tr}$
\ENDIF
\ELSE
\STATE set $T_{det}(xcur, \xi, \gamma)$ = 1 (index to failure state)
\ENDIF
\ENDFOR
\ENDFOR
\STATE cur $\leftarrow$ cur + 1
\ENDWHILE
\RETURN S, $T_{det}$
\end{algorithmic}
\end{algorithm}
Once we have the deterministic state transition matrix, we proceed to calculate the stochastic state transition matrix and other important elements needed to perform our analyses as explained in the following sections. 
\subsection{Stochastic State Transition matrix}
To analyze the performance of a particular DRL-based control policy, $\xi$, when subject to some disturbance profile, $P(\gamma)$, we first calculate a \textit{stochastic} transition matrix, $T(i,j)$:
\begin{equation}
\label{eq:Ts}
T(i,j):=\sum_{\gamma}P(\gamma)f_j
\end{equation}
where
\begin{equation}
\label{eq:Ts_cond}
f_j =
  \begin{cases}
    1,  & \quad \text{if } T_{det}(i,\pi(i),\gamma)=j\\
    0,  & \quad \text{else}.
  \end{cases}
\end{equation}
Matrix $T(i, j)$ gives the probability of transitioning from state $i$ to state $j$ within mesh $\mathbb{M}$. Because all states must transition to some state under the entire probability distribution of the disturbance profile, each row of the stochastic state transition matrix sums to 1.  

\subsection{Mean First Passage Time}
The concept of the mean first passage time (MFPT) for metastable (i.e., ``rarely falling'') walking systems is explained in detail in \cite{Oguz2015} and \cite{Byl2009MetastableWM}. We highlight important results here that we will be using in the rest of the paper. To analyze not only convergence rates (eigenvalues) but also probability distributions across state space (via eigenvectors), we focus on an eigenanalysis of the \textit{transpose}, $T^T$, of the stochastic transition matrix, $T$. First, we define a metastable distribution, $\phi$, as the stationary distribution in state space, conditioned on not having entered the failure state:
\begin{equation}
\label{eq:stationary}
\phi_i = \lim_{n\rightarrow \infty}Pr(X(n)=x_i | X(n)\neq x_1).
\end{equation}
The mean first passage vector $m$, where $m_i$ is the mean time for the state $i$ to go into failure, is given by
\begin{equation}
\label{eq:MFPT_vector}
m_i  =
\begin{cases}
0, & \quad \text{if } i = 1\\
1 + \sum_{j > 1}T_{ij}m_j, & \quad \text{else}.
\end{cases}
\end{equation}
Vector $m$ can be calculated by
\begin{equation}
\label{eq:MFPT_matrix}
m = 
\left[
\begin{tabular}{c}
0  \\
$(I-\hat{T})^{-1}$
\end{tabular}
\right],
\end{equation}
where $\hat{T}$ is $T$ with the first column and row (corresponding to the failure state, for which $m(1)=0$) removed. This in turn lets us calculate the system wide MFPT:
\begin{equation}
\label{eq:MFPT}
 M=\sum_i \phi_im_i.
\end{equation}
This process can be time consuming and for a more efficient calculation of $M$, we use the eigenvalues of the stochastic transition matrix. As stated, all failures (e.g., falls) are modeled via a self-absorbing state. This enforces a structure on the stochastic state transition matrix where $T(1,1)=1$, resulting in the largest eigenvalue of $T$ being 1. Let $\lambda_2$ be the second largest eigenvalue of $T$. The magnitude of $\lambda_2$ is the probability of taking a successful step, assuming initial conditions have been forgotten and the walker is in a non-failure state; i.e., that the likelihood of currently being in any state is given by the metastable distribution, $\phi$. The probability of failure on the next step is then $1 - \lambda_2$, and the probability of taking only $n$ steps is
\begin{equation}
\label{eq:n_steps}
Pr(X(n)=x_1, X(n-1)\neq x_1) = \lambda_2^{n-1}(1-\lambda_2).
\end{equation}
When the dynamics of $\lambda_2$ are much slower than all smaller-magnitude eigenvalues, i.e., when $(1-|\lambda_2|)<<(1-|\lambda_3|)$, initial conditions will be forgotten long before an expected failure event, and a system-wide mean first passage time, $M$, provides a useful metric for stability:
\begin{equation}
\label{eq:MFPT_approx}
M = \frac{1}{(1-\lambda_2)}.
\end{equation}
Finally, note that normalizing all the non-failure-state elements in the eigenvector associated with the $2^{nd}$-largest eigenvalue of $T^T$ (the \textit{transpose} of $T$) correspondingly yields the metastable distribution, $\phi$, in Eq.~\ref{eq:stationary}.

\section{RESULTS}
\label{sec:results}
We perform analyses on two DRL-trained policies, obtained under different training conditions. The first policy is trained with a reward function that encourages forward velocity; there are no perturbations, and terrain is flat. At each time step, if the walker has not fallen, the reward is incremented by the forward velocity at that instant of time; the reward function also includes a penalty 1e-3 times the norm of the torques. For the second scenario, we use the same reward function, but we now introduce push disturbances while training (still on flat terrain). For both cases, we train several policies and then pick the one that has the maximum reward at the end of the training session for our analyses. Both cases also have the same coefficient of friction of 0.5 for contact between the ground and the walker model. For meshing, we ignore the $x$ coordinate of the top of the torso, because we perform meshing for step-to-step transitions and it does not matter from what $x$ position the model takes the step. The mesh thus contains 13 dimensional states, of which 12 are independent. (The left foot, but not necessarily the right one, must by definition be at $y=0$ immediately post-impact, adding a constraint and removing an independent DOF on the Poincar\'e section.) We explore both cases and analyze the corresponding policies obtained in the next two subsections.

\subsection{Case 1}
\label{subsec:scenario 1}
\begin{figure}[!ht]
\centering
\includegraphics[width = 3.3 in]{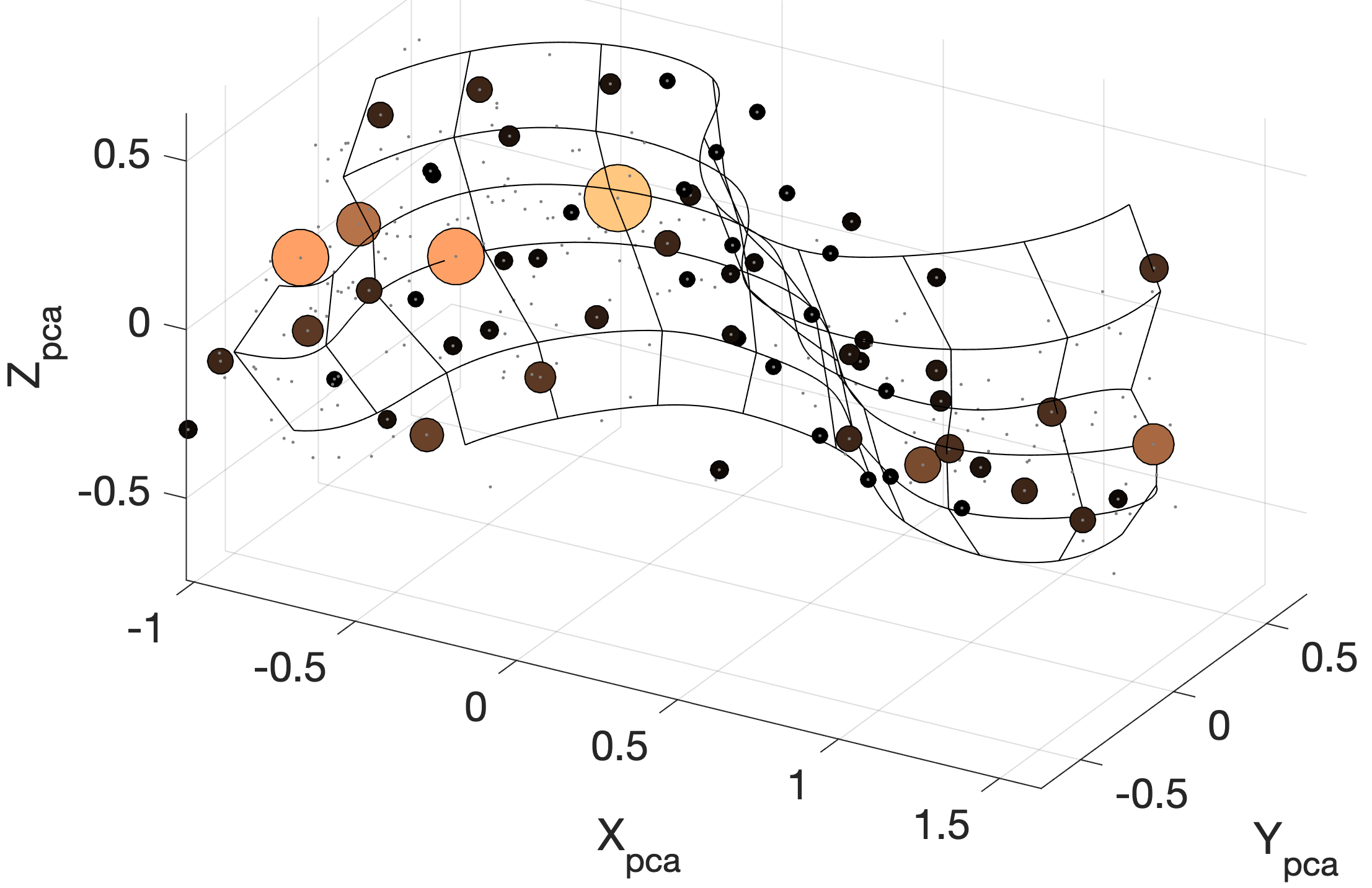}
\includegraphics[width = 3.3 in]{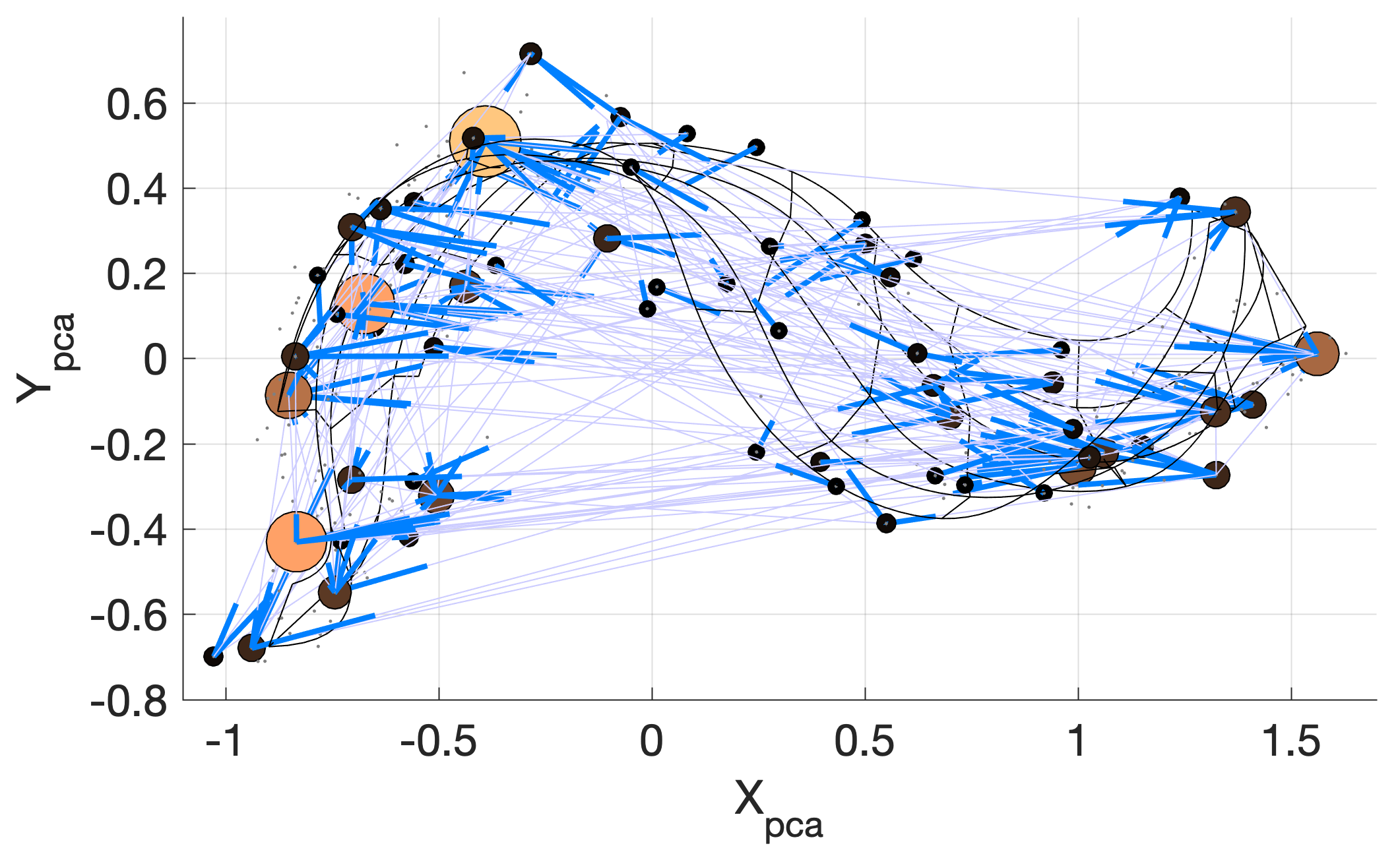}
\caption{A principal component analysis (PCA) is used to visualize post-impact states visited for DRL-trained flat-ground locomotion, subject to no disturbances. See text for a detailed explanation.}
\label{fig:chaotic_gait}
\end{figure}

In this scenario, we train the policy without any external perturbations, for flat ground walking. The policy with the best reward function, chosen for our analyses, results in a significantly asymmetric forward motion. As previously mentioned, we consider state-to-state transitions for a full gait cycle, defined as one complete sequence of right leg and subsequent left leg contacts with the ground. The states on the Poincar\'e section then represent the post impact state of the system each time the left leg makes contact with the ground. Fig.~\ref{fig:chaotic_gait} illustrates a typical set of consecutive, post-impact states during locomotion (again, with no perturbations) for 250 gait cycles.

Figure~\ref{fig:chaotic_gait} shows a non-uniform mesh of 64 points, achieved by applying our meshing algorithm to a set of 250 consecutive gait cycles, for the purposes of giving more intuition both into our meshing an into the lower-dimensional nature of these points. Here, sufficiently close neighbors within the full set of 250 states visited are ``lumped together'' at one of 64 total mesh points. Larger markers depict more frequently visited locations. For this particular figure, we performed a PCA analysis, to generate 3D axes to visualize the mesh. Here, three principle components account for $94\%$ of the variance of the normalized Poincar\'e states. At top in Fig.~\ref{fig:chaotic_gait}, states fall near the depicted curvy, 2D surface within the 3D PCA space. Below this, straight lines show the step-to-step transitions, with terminal ends shown as thicker, blue segments. The gait has no observable limit cycle, yet it appears to be both stable (never-falling)~-- and chaotic.

As Figure~\ref{fig:chaotic_gait} illustrates, applying the closed-loop DRL control policy to flat-ground walking with no noise inputs does not result in any discernable (period-n) limit cycle but instead exhibits chaotic behavior. Based on our numeric methods for mapping the reachable state space, post-impact states are bounded and seem to contract onto a lower-dimensional and bounded region of the full state space.

To examine the robustness of the given policy to push disturbances, we proceed by calculating a mesh. 
From simulations, we find the complete two-step gait cycle takes about 0.5~(s). Based on this, for building the mesh, we define a threshold of 0.3~(s) after which any post impact (for the left foot) state of the system, $s_i$, is added to the mesh if $d(s_i)>d_{tr}$, as described in Section~\ref{sec:mesh}. 


For illustrative purposes within Case~1, just four possible disturbance pushes are considered: a push of either $+1000$ (N) or $-1000$ (N) is applied at the COM of the torso for a duration of 0.008~(s), starting at either 0.1 or 0.25~(s) into the gait cycle.
Along with these 4 disturbances, we also consider a no-push disturbance, i.e., a push with 0 magnitude. The probability distribution for these disturbances is given by
\begin{equation}
\label{eq:P_distribution}
P(\gamma)  = 
\begin{cases}
0.4, & \quad \text{if no disturbance}\\
0.6/4, & \quad \text{else}.
\end{cases}
\end{equation}
For the given policy and disturbance profile, we obtain a mesh with 28,757 states when $d_{tr} = 0.6$. The mean first passage time for the system for the given probability distribution comes out to only about 32 steps. Once the mesh is generated, our tools allow us to efficiently calculate the MFPT for any given probability distribution. For example, if we reduce the probability of a disturbance so that
\begin{equation}
\label{eq:P_distribution_eg}
P(\gamma)  = 
\begin{cases}
0.8, & \quad \text{if no disturbance}\\
0.2/4, & \quad \text{else},
\end{cases}
\end{equation}
the MFPT increases to approximately 117 steps. 
We also calculate a mesh for other values of $d_{tr}$, to analyze how varying this threshold distance changes the number of states in the mesh. If the mesh points occupy an n-dimensional subspace within the full state space, then the number of mesh points, $N$, required to span this subspace should grow as $N \propto d_{tr}^{-n}$, meaning a loglog plot of $d_{tr}$ vs $N$ would have slope $-n$. (See \cite{Meshing_nihar_2019} for details.) 

\begin{table}[ht]
\begin{center}
\begin{tabular}{|c|c|c|c|c|}
\hline
$d_{tr}$ & 0.6 & 0.7 & 0.8 \\
\hline
$N$ (\# of mesh points) & 28,757 & 14,891 & 8,517 \\
\hline
\end{tabular}
\end{center}
\caption{$N$ versus $d_{tr}$, showing $n \approx 4.23$ for Case 1.}
\label{tab:N_points_C1}
\end{table}
\vspace*{-.2in}
Table~\ref{tab:N_points_C1} shows this variation. A line fitting $x=log(d_{tr})$ vs $y=log(N)$ for these data has slope $-4.2$, showing that as we mesh more finely, the size of the mesh grows with dimensionality $n \approx 4.23$. 
Intuitively, the dimensionality of reachable state space will depend on both the sparsity of the perturbation set (recall we only include 4 non-zero perturbations here) as well as the contracting nature of the control policy.  Rather than focusing on a more dense set of perturbations for Case 1, we instead focus on comparing the effects of adding perturbations during training (in Case 2).

\begin{figure}[!ht]
\centering
\includegraphics[height = 1.25 in]{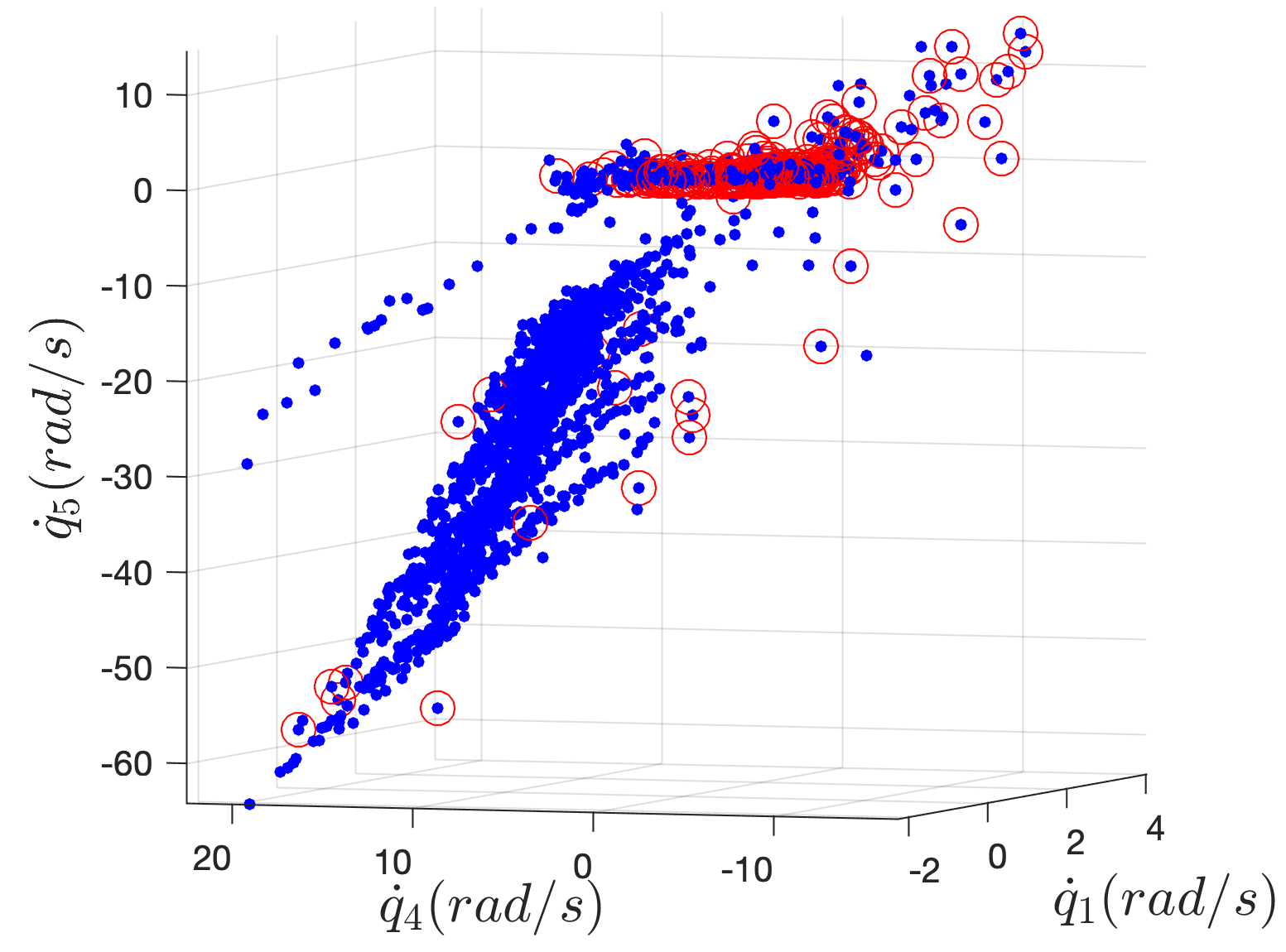}~~~~
\includegraphics[height = 1.25 in]{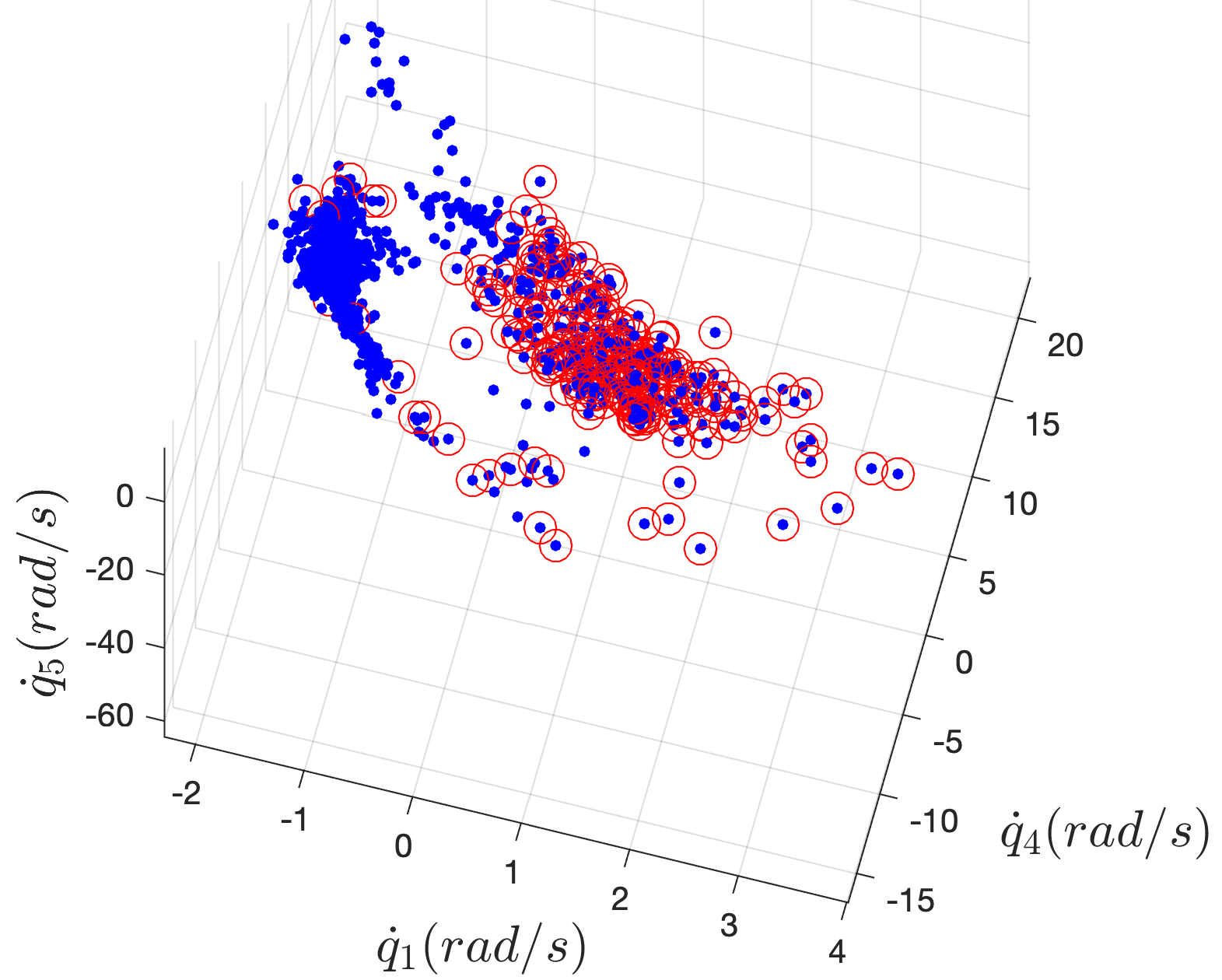}
\caption{A 3D section of the full 13D Poincar\'e mesh states for Case~1. Subplots show two viewpoints of the same data. ``Dangerous'' states, with greater than $99\%$ probability of failure on the next step, are circled in red.}
\label{fig:case1_states}
\end{figure}
Figure~\ref{fig:case1_states} illustrates the distribution of mesh points for Case 1, using states with the greatest variance within the mesh ($\dot{q}_4$, $\dot{q}_5$, and $\dot{q}_1$) as three representative axes. Recall that even without added perturbations, the control policy results in a chaotic set of reachable states; these points fall within the blue region. 
The red points show states visited due to perturbations and from which failures are nearly certain. 

\subsection{Case 2}
\label{subsec:scenario 2}
In this scenario, we include push disturbances during the training, again using a reward function that primarily encourages forward motion, with a small penalty on energy use. Every 0.008~(s) interval during training, there is a 5 percent chance that the model will receive a push for the next 0.008~(s), with a 50/50 chance in either the forward or the back ward direction. We only introduce pushes at the COM of torso, and the magnitudes of the push are restricted to $\pm$1000~(N). 
As in Case 1, the policy with the maximum reward that we obtain for this new scenario also results in an asymmetric gait, now with a slower average gait cycle time of approximately 0.5 (s). We pick a threshold time of 0.45 (s) after which any post impact state (for left foot) with the ground is considered for meshing.
To examine the effects of introducing disturbances during the training, we generate a mesh for the same disturbance profile (4 possible pushes, plus one no-push case) used in Case 1; for $d_{tr}=0.6$, this leads to a mesh of 857 states. For the probability distribution given in Eq.~\ref{eq:P_distribution}, the MFPT for the system now comes out to about 10,467 steps, as compared with 32 steps for Case 1. An improvement in performance is an intuitively expected result, but our focus is on illustrating that our meshing tools allow us to quantify the improvement in robustness. 
We also generate the mesh for different $d_{tr}$ values, shown in Table \ref{tab:N_points_C2}. 

\begin{table}[ht]
\begin{center}
\begin{tabular}{|c|c|c|c|c|}
\hline
$d_{tr}$ & 0.5 & 0.6 & 0.7 \\
\hline
$N$ (\# of mesh points) & 1705 & 857 & 574 \\
\hline
\end{tabular}
\end{center}
\caption{$N$ versus $d_{tr}$, showing $n \approx 3.25$ for Case 2.}
\label{tab:N_points_C2}
\end{table}
\vspace*{-.2in}

\begin{figure}[!ht]
\centering
\includegraphics[width = 3.0 in]{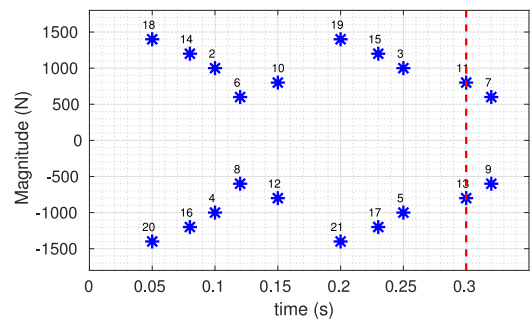}
\caption{Disturbance profile for 20 push types. The dotted red line indicates the half gait cycle where the right leg makes contact with the ground}
\label{fig:disturbance_profile}
\end{figure}

To analyze a more interesting disturbance profile and to check how the policy performs for magnitudes beyond the ones used for training, we generate a more complex mesh involving more varied disturbances, as shown in Fig.~\ref{fig:disturbance_profile}
As with the previous mesh, in addition to the 20 disturbances shown, we also include a no push scenario. This mesh has a total of 20,660 states which is significantly higher than the previous mesh but because we explored the mesh for more disturbances, it also allows us to do much more interesting analyses. For example, if we set the probability distribution of the disturbances such that there is 0.6 probability (60\% chance) of no push and 20\% chance for disturbance 6 and disturbance 7 each, then we get a MFPT of infinity indicating that the walker will always recover under such disturbances. Similarly, if we set the probability disturbance such that probability of no push is 0.6, and probability of disturbance 8 and 9 is 0.2 each, then we get a MFPT of infinity as well. But, if we set the distribution such that probability of no push is 0.6 and the probability of disturbance 6, 7, 8 and 9 is 0.1 each, we get an MFPT of 32,260 showing that the \textit{mixing effects} of disturbance 6, 7 and disturbance 8, 9 all combined over time will now create occasional failures, reducing the MFPT of the system significantly. Disturbances 6 and 7 correspond to pushes in the forward direction of magnitude 600 (N) and disturbance 8 and 9 correspond to the backward pushes of the same magnitude. We can also study the relative sensitivity of particular disturbances in the profile. For this we take an example distribution given by: 
\begin{equation}
\label{eq:P_distribution_MFPT}
P(\gamma)  = 
\begin{cases}
0.4, & \quad \text{no disturbance}\\
0.5, & \quad \text{disturbance of interest}\\
0.1/19, & \quad \text{else}.
\end{cases}
\end{equation}

\begin{figure}[!ht]
\centering
\includegraphics[width = 3 in]{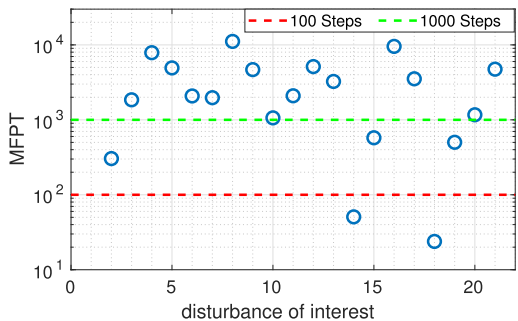}
\caption{MFPT variations as one disturbance of interest becomes more likely. The MFPT is shown on log scale to make the plot more readable.}
\label{fig:disturbance_MFPT}
\end{figure}

A plot showing the corresponding performance of the DRL policy for Case 2 is shown in Fig.~\ref{fig:disturbance_MFPT}.
This is an interesting plot because it shows the coupled effect of direction and time of impact of the disturbance matters significantly. For example, we see that some disturbances of higher magnitude have a better MFPT than some disturbances of lower magnitude because they happen at different time instances. 

\vspace*{-.1in}

\begin{figure}[!ht]
\centering
\includegraphics[height = 1.35 in]{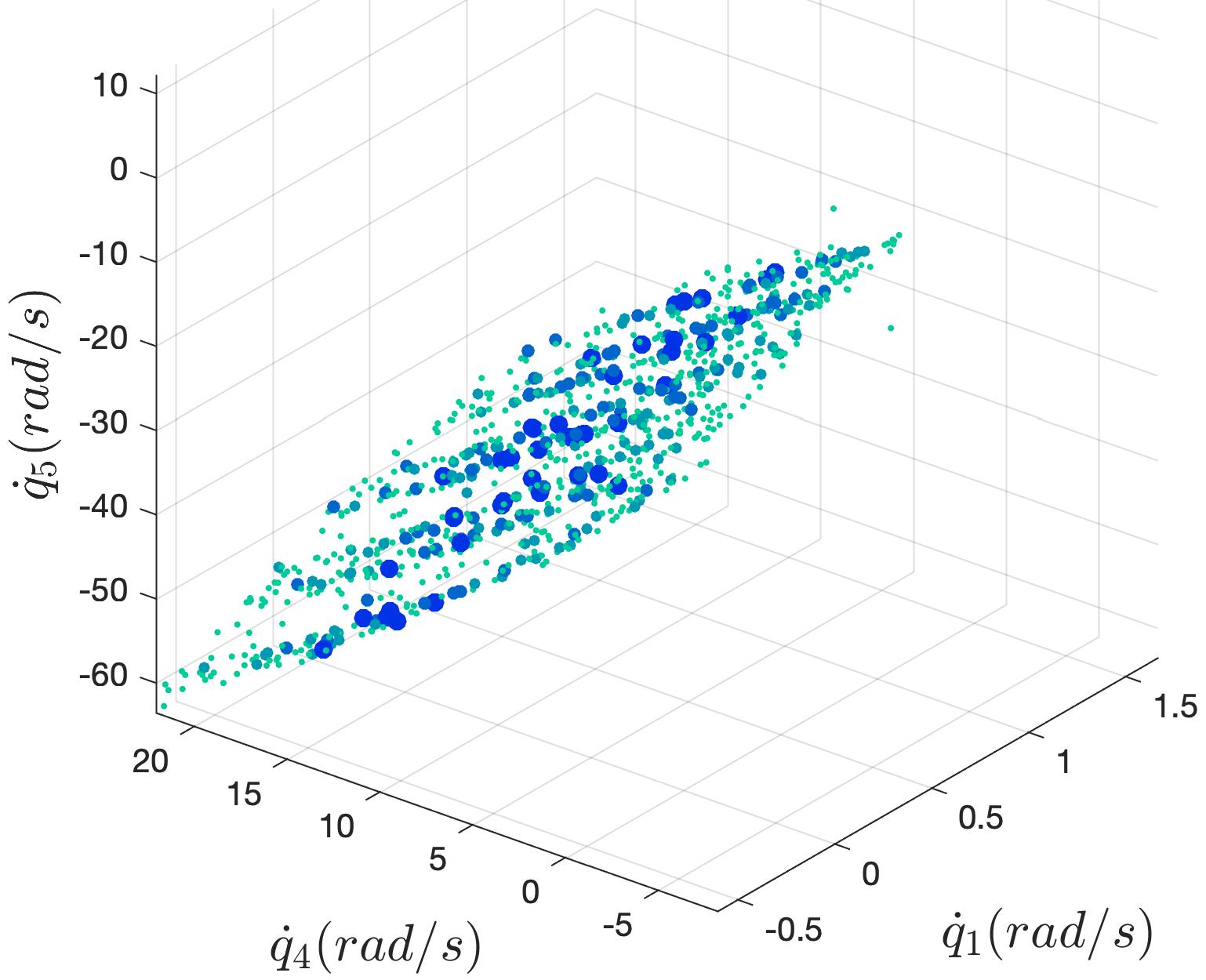}\includegraphics[height = 1.35 in]{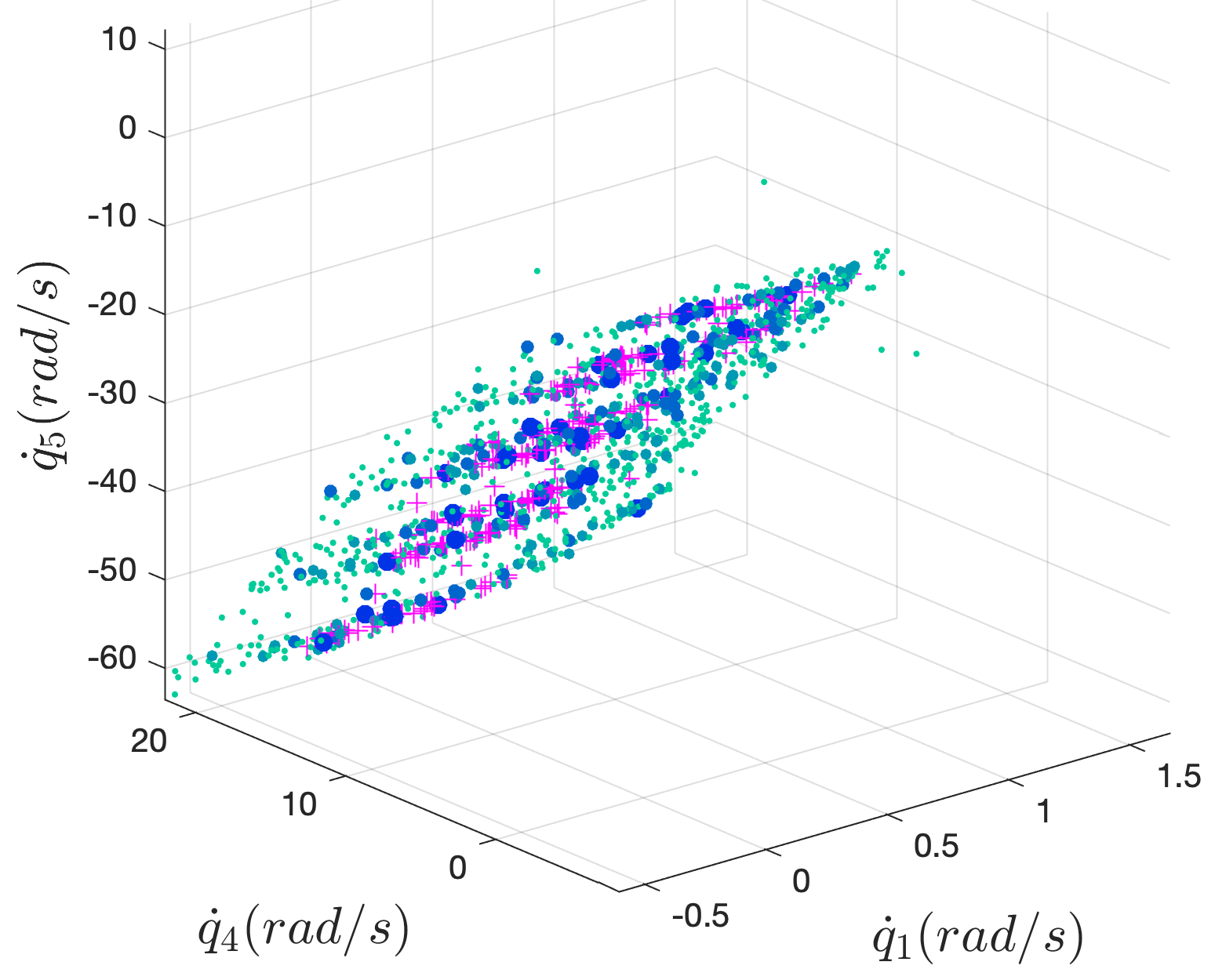}
\caption{A 3D slice of the full 13D Poincar\'e states generated to mesh the policy trained in Case~2, when all but disturbance profiles 14, 16, 18 and 20 are included, i.e., excluding pushes occurring before 0.1~seconds. }
\label{fig:case2_states_caseKB1}
\end{figure}

\begin{figure}[!ht]
\centering
\includegraphics[height = 1.38 in]{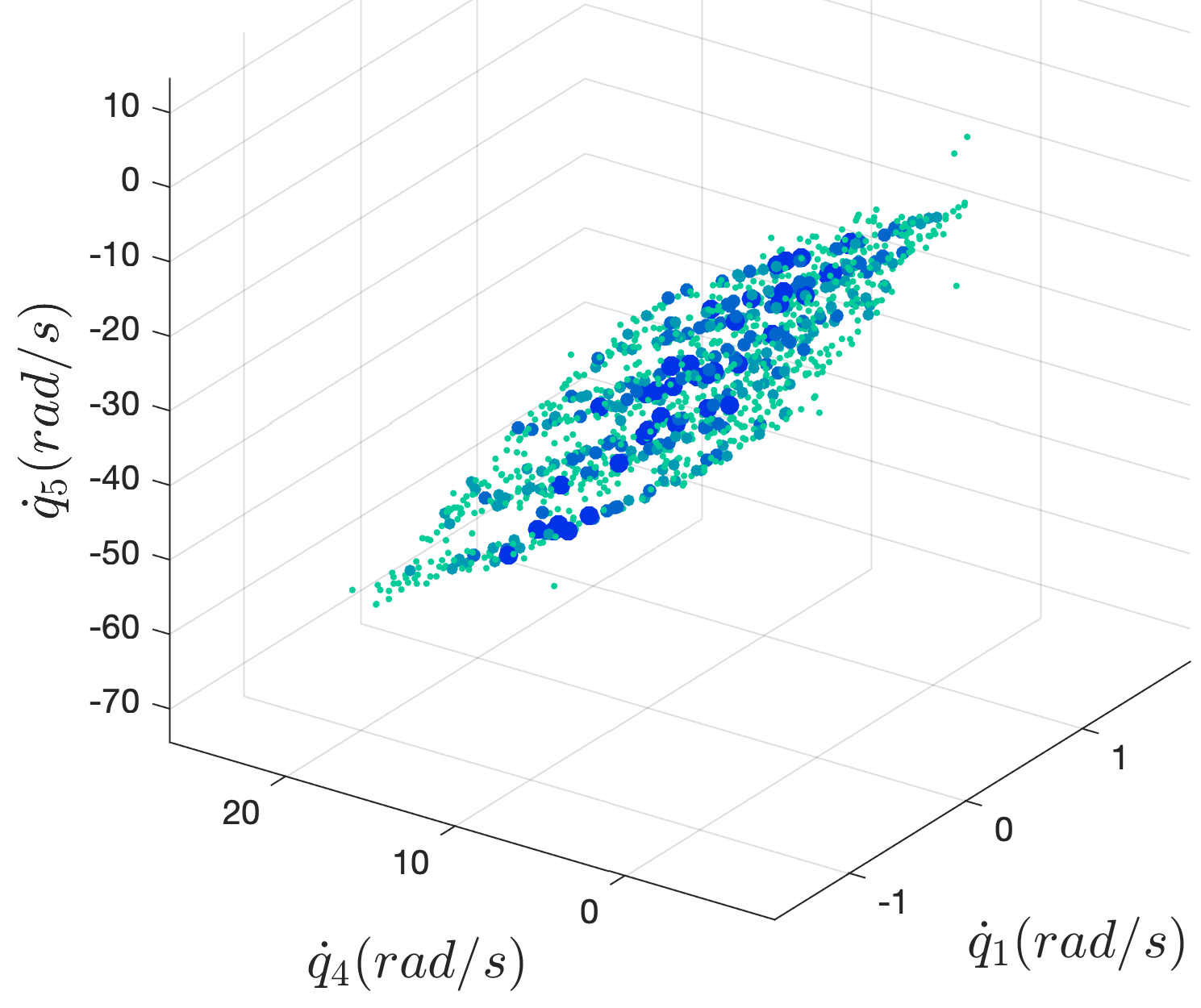}\includegraphics[height = 1.38 in]{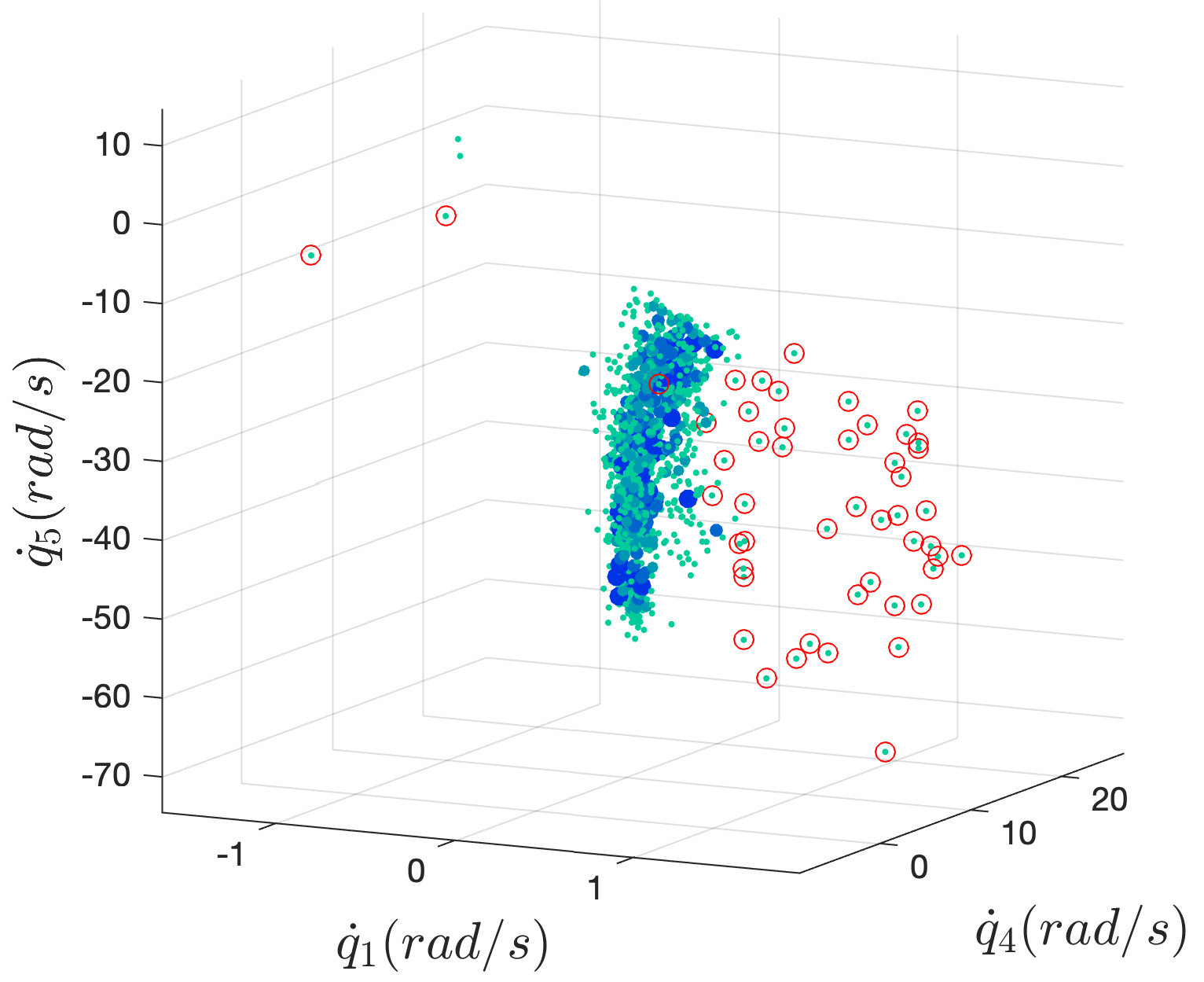}
\caption{A 3D slice of the full 13D Poincar\'e states generated to mesh the policy trained in Case~2, when all 20 timing and magnitude combinations shown in Fig.~\ref{fig:disturbance_profile} are included. As in Fig.~\ref{fig:case1_states}, the subplot at right highlights states from which immediate failure has probability greater that $99\%$.}
\label{fig:case2_states_caseKB2}
\end{figure}

Figures~\ref{fig:case2_states_caseKB1} and \ref{fig:case2_states_caseKB2} show subsets of the full mesh corresponding to the reachable state space when each of two different subsets of the disturbance profiles shown in Fig.~\ref{fig:disturbance_profile} are allowed.  For Fig.~\ref{fig:case2_states_caseKB1}, we exclude any pushes occurring before 0.1 seconds in the gait cycle; i.e., profiles 14, 16, 18 and 20 are excluded. For Fig.~\ref{fig:case2_states_caseKB2}, all 20 push disturbances are included.  For both figures, we assume a $2\%$ chance of a push during each gait cycle, with the push type drawn with uniform probability from the allowable subset of pushes.

In Fig.~\ref{fig:case2_states_caseKB1}, both subplots show the mesh points visited for this noise case, with darker points representing much more frequently-visited states. At right, magenta ``+'' symbols are overlaid to show a set of 250 consecutive states visited (chaotically) when there is no noise during post-training testing of the policy. The MFPT predicted by the mesh is about 958,000 gait cycles. Once all 20 pushes are allowed (Fig.~\ref{fig:case2_states_caseKB2}), this drops to a MFPT of only about 4,900 gait cycles. In this latter case, we can see that the system now visits a significant number of ``unsafe'' states (shown in red) that department significantly from the chaotic variability of locomotion when there is no noise. 

\vspace*{-.1in}

\section{CONCLUSIONS}
\label{sec:conclusion}
We have demonstrated the effectiveness of our meshing tools in analyses of the polices created via deep reinforcement learning. To quantify the robustness of these policies, we create meshes for different disturbance combinations that can happen at various times in the gait cycle and then set transitions among mesh points based on particular assumed probability distributions of the disturbances. The policy trained for only forward motion is significantly less robust to perturbations as compared to the policy trained with impacts, which is intuitive but our tools gave us an estimate on the actual performance improvement. Such an estimate is important if we want to improve the performance of training by changing various parameters that can affect the outcome of the trained policy. In addition, by performing the analyses on various probability distributions we show that our tools also allow us to  plot performance trends which can help us to understand the effect each individual disturbance in the disturbance profile has on the overall performance of the policy. We also show how the mixing effects of these disturbances can have significant effect on robustness. 

In showing the applicability of our meshing tools to analyze the policies obtained via deep reinforcement learning, we have also created an indirect feedback loop that can be used to improve the performance of the policies by tuning various elements of the training framework. A future goal is to integrate this feedback loop within the training framework. We also plan to explore how increasing the number of discrete perturbation types affects mesh size. Although we consider toy sets of perturbations here, for illustrative purposes, practical use of these tools should cope with denser sets of perturbations. 
Finally, another important goal is to explore how various reward functions and/or disturbances during training may increase the degree to which a DRL policy contracts the dynamics; reducing the dimensionality of the state space visited would increase the practicality of meshing. We show some initial success in this by reducing dimensionality from $n=4.23$ to $n=3.25$ when training includes pushes. Future work is planned to investigate if this a repeatable trend and whether more extreme perturbations during training can further improve contraction. 








.

\bibliographystyle{IEEEtran}
\bibliography{CDC2019}

\end{document}